\documentclass[runningheads]{llncs}

\usepackage[mobile,year=2024,ID=7954]{eccv}

\usepackage{eccvabbrv}

\usepackage{graphicx}
\usepackage{booktabs}

\usepackage[accsupp]{axessibility}  
\usepackage[dvipsnames]{xcolor} 

\usepackage{amsmath}

\DeclareMathOperator*{\argmin}{arg\,min}

\usepackage[pagebackref,breaklinks,colorlinks]{hyperref}

\usepackage{orcidlink}
\usepackage{multirow}

\begin{document}

\newcommand\aalok[1]{\noindent\color{RawSienna}{#1}}
\newcommand\callum[1]{\noindent\textcolor{violet}{#1}}
\newcommand\gwangbin[1]{\noindent\textcolor{blue}{#1}}
\newcommand\tildetext{{\raise.17ex\hbox{$\scriptstyle\sim$}}}

\title{U--ARE--ME: Uncertainty-Aware Rotation Estimation in Manhattan Environments} 

\titlerunning{U--ARE--ME}

\author{Aalok Patwardhan \and
Callum Rhodes \and
Gwangbin Bae \and
Andrew J. Davison}

\authorrunning{A. Patwardhan et al.}

\institute{Dyson Robotics Lab, Imperial College London 
\email{\{a.patwardhan21, c.rhodes, g.bae, a.davison\}@imperial.ac.uk}} 

\maketitle

\begin{figure}[!h]
    \centering
    \includegraphics[width=0.8\columnwidth]{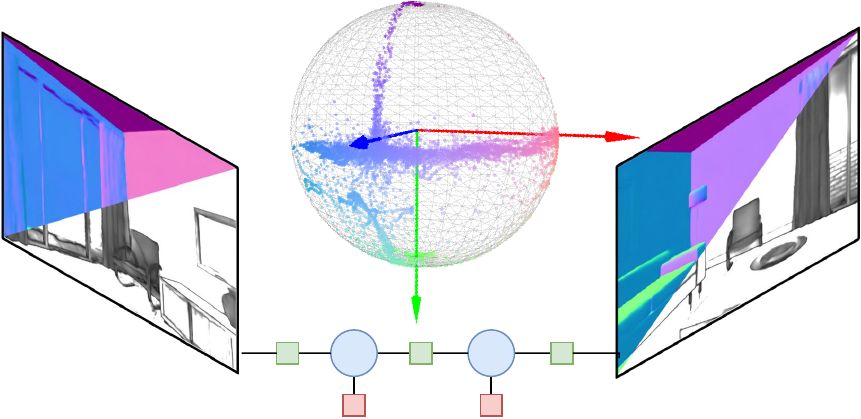}
    \caption{U--ARE--ME provides globally consistent rotation estimates in Manhattan environments across sequences of RGB images -- without camera intrinsics. For each frame, we estimate the rotation from the predicted surface normals along with pixel-wise uncertainty and enforce temporal consistency via a factor graph.}
    \label{fig:enter-label}
\end{figure}

\vspace{-1cm}
\begin{abstract} 
Camera rotation estimation from a single image is a challenging task, often requiring depth data and/or camera intrinsics, which are generally not available for in-the-wild videos. Although external sensors such as inertial measurement units (IMUs) can help, they often suffer from drift and are not applicable in non-inertial reference frames.
We present U-ARE-ME, an algorithm that estimates camera rotation along with uncertainty from uncalibrated RGB images. Using a Manhattan World assumption, our method leverages the per-pixel geometric priors encoded in single-image surface normal predictions and performs optimisation over the SO(3) manifold.
Given a sequence of images, we can use the per-frame rotation estimates and their uncertainty to perform multi-frame optimisation, achieving robustness and temporal consistency.
Our experiments demonstrate that U-ARE-ME performs comparably to RGB-D methods and is more robust than sparse feature-based SLAM methods.
We encourage the reader to view the accompanying video at https://callum-rhodes.github.io/U-ARE-ME for a visual overview of our method.

\end{abstract}

\section{Introduction}
\label{sec:intro}
\begin{figure}[t]
    \centering
    \includegraphics[width=1.0\linewidth]{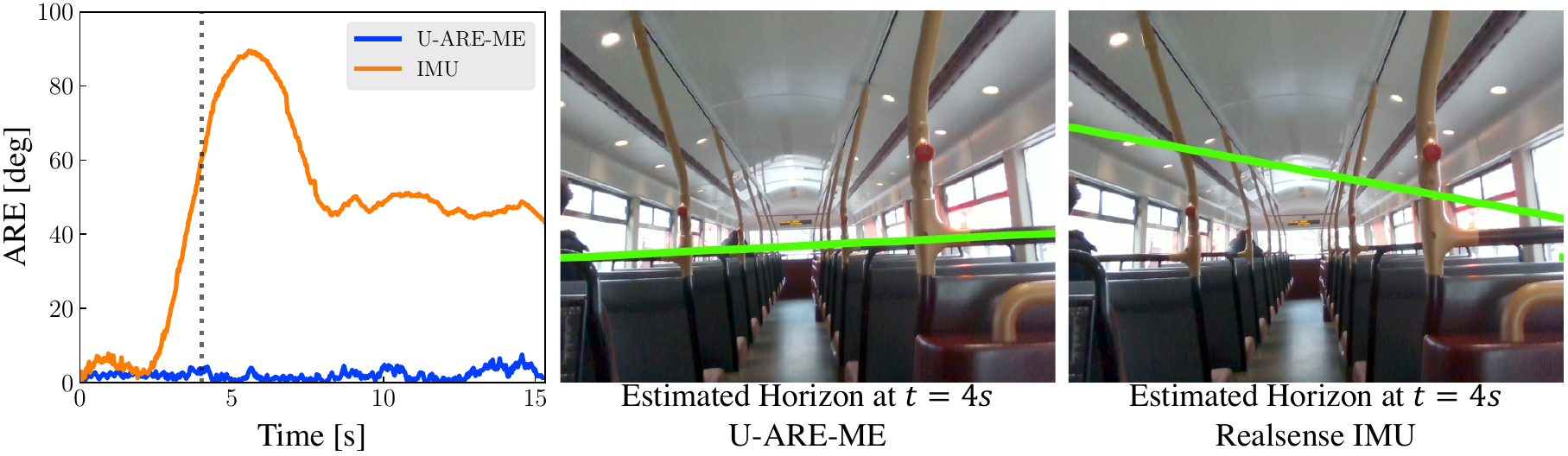}
    \caption{IMU sensors are prone to drift especially in non-inertial frames of reference (e.g. inside a moving vehicle). The horizon line in each image represents the line perpendicular to the \textit{up-vector} inferred from our proposed method (middle) and an IMU sensor (right).}
    \label{fig:IMU_horizon}
\end{figure}

Accurate estimation of camera rotation from a sequence of monocular images is crucial for many computer vision applications, including visual odometry~\cite{he2020review}, image stabilisation~\cite{morimoto1998evaluation}, and augmented reality~\cite{augreality}. Many solutions have been proposed for a variety of sensor setups. For instance, the recently released Apple Vision Pro operates using visual-inertial odometry, relying on both the cameras and the inertial measurement units (IMUs). However, IMUs are prone to drift and are by design not suitable for non-inertial frames of reference (see Fig.~\ref{fig:IMU_horizon}).

If depth measurements (paired with the input images) are available, the RGB-D frames can be aligned --- based on photometric and geometric consistency --- to recover their relative camera poses~\cite{bundlefusion}. If the surface normal vectors in the scene are aligned with a set of \textit{principal directions}, the camera rotation can be found by aligning the input normals (extracted from the depth maps) to those directions~\cite{SilbermanES,Ghanem,StraubRTMF}. While such approaches provide drift-free rotation estimates with high accuracy, they cannot be applied to in-the-wild videos or devices without a depth sensor.

This paper focuses on the most challenging setup in which only RGB input is available. Previous attempts have focused on detecting and matching \textit{2D image features}. For instance, ORB-SLAM \cite{mur2015orb} tracks sparse ORB features, while methods like \cite{olre, Lee-ROVE} group line segments to identify the vanishing points and hence the camera rotation with respect to the principal directions. However, such methods are sensitive to image degradation (e.g. noise and motion blur) and perform poorly in textureless environments. More importantly, they assume known camera intrinsics --- which are often not available for in-the-wild videos. While a neural network can be trained to regress the rotation between consecutive frames~\cite{cai2021extreme}, such an approach is prone to overfitting and drift. It is also computationally costly to train such a specialised model.

In this work, we propose to make use of the dense pixel-wise geometric priors learned by \textit{single-image surface normal estimation models}. Surface normal estimation models are efficient (e.g. \cite{SNfromRGB_2021_EESNU} runs at \tildetext60 fps on a 4090 GPU) and have strong generalisation ability~\cite{SNfromRGB_2021_EESNU,bae2024dsine}. In recent years, their usefulness has been demonstrated for various computer vision tasks, including object grasping~\cite{zhai2023monograspnet}, multi-task learning~\cite{liu2023prismer}, simultaneous localisation and mapping~\cite{li2020structure}, and CAD model alignment~\cite{langer2022sparc}. We explore whether such powerful front-end perception can also be used for rotation estimation.

Similar to previous optimisation-based approaches~\cite{SilbermanES, Ghanem, StraubRTMF}, we assume a certain distribution of surface normal vectors in world coordinates and optimise for the camera rotation that would align the predicted normals to the principal directions of the scene. While previous methods (1) used depth sensors to extract the normal vectors and (2) were only applicable to a single image, we attempt to remove both constraints. Two types of uncertainty arise in the process of removing these two commonly adopted constraints.

First is the heteroscedastic aleatoric uncertainty~\cite{UNC_2017_what_do_we_need} in surface normal predictions. As shown in \cite{SNfromRGB_2021_EESNU}, surface normals predicted by a neural network --- unlike those extracted from a depth map --- are unreliable, especially for the pixels near object boundaries and on small objects. As these pixels should be down-weighted in the optimisation objective, we introduce a new uncertainty-weighted cost function and show how the uncertainty can be learned in a data-driven manner.

The second type of uncertainty arises when the image contains a limited number of principal directions. For instance, when a Manhattan World (MW) ~\cite{MWVP_2000_MW} is assumed, two (or more) of the six directions ($\pm X, \pm Y, \pm Z$) should be observed to determine the camera rotation. If only one axis is visible, any rotation around that axis would result in an equally valid prediction. To this end, we quantify the uncertainty around each principal axis and use it to enhance the temporal consistency in the predictions.

To summarise, our framework alternates between two optimisation steps:

\begin{itemize}
    \item \textbf{Single-frame optimisation:} We optimise the world-to-camera rotation matrix such that the rotated principal directions are best aligned with the predicted surface normals. We improve the accuracy and robustness by introducing an uncertainty-weighted cost function.
    \item \textbf{Multi-frame optimisation:} We take the covariance matrix of rotation around each axis --- which is readily available from the Hessian approximation in the second-order optimisation of the first step --- and use it to jointly optimise a sliding window of previous frame rotations. We improve the global consistency of our solution, reject outlier rotations and intuitively handle frames that may contain limited information on certain principal axes.
\end{itemize}

The proposed method runs at \tildetext40 fps on an NVIDIA 4090 GPU. Note that, unlike the learning-based models that can only be used for rotation estimation, the surface normal predictions --- from which we infer the rotation --- can be used for other tasks, reducing the overall computational overhead.

The main strength of our approach lies in its \textit{robustness}. Compared to the methods that rely on sparse feature tracking or line segment detection, our approach is more robust to the presence of image degradation. Unlike SLAM-based methods, our approach does not require camera intrinsics and can be applied to a single image or in-the-wild videos, making it useful in a wider range of scenarios.

\section{Related work}
\label{sec:related}

Rotation estimation of a camera from single images has been extensively studied and is generally based upon the assumption that indoor scenes exhibit inherent structure, conforming to the MW assumption. Estimating the Manhattan Frame (MF) -- the three principal orthogonal directions of the MW scene thus becomes the primary focus of rotation estimation approaches. These approaches for MF estimation broadly fall into two domains; using RGB images to extract perspective cues, and using 3D information such as surface normals from RGB-D images.

Earlier work estimates the MF by considering vanishing points and lines in an image as perspective cues \cite{VideoCompass}. The work in \cite{Lee-2009} generates several MF hypotheses from an image, using line segments to find the best fitting model. In \cite{Furukawa} line segments are extracted onto a hemisphere and clustered to identify three orthogonal directions, however, this method is sensitive to the chosen resolution of the hemisphere discretisation.
The algorithm proposed by \cite{Elloumi} uses line clustering to find three vanishing points, and achieves real-time camera rotation estimation over a sequence of images in a video. The work in \cite{Lee-ROVE} (ROVE) does not rely on the MW assumption but uses sequential Bayesian filtering to jointly estimate rotation and vanishing points.
These RGB methods rely on the existence of multiple parallel lines and vanishing points, and are not robust in the presence of noise and outliers, or in texture-less scenes. 

Rotation estimation methods that use RGB-D images are more accurate and stable as they utilise 3D information in the scene, whether this is directly through depth camera data, or by using this data to compute the surface normals in the scene.
The approach in \cite{SilbermanES} uses point normals and perspective cues to perform an Exhaustive Search (ES) over a set of candidate directions and uses a scoring heuristic to estimate the MF, although this incurs a high computational cost.
Depth data from a Kinect camera was used in \cite{TaylorKinect} to determine the MF of a scene by identifying the ground, and selecting a perpendicular direction from one of the walls. This method relies on the presence of a visible floor and multiple walls in the image, and so is not generalisable.
In \cite{Ghanem} the MF is estimated through non-convex optimisation by considering the sparsity constraints of MF-aligned surface normals.

In \cite{StraubMMF} the authors argue that real-world scenes contain a Mixture of Manhattan Frames (MMF), which they simultaneously estimate from surface normals calculated from depth data. Their follow-up work -- RTMF~\cite{StraubRTMF} -- presented a GPU-accelerated version for real-time applications.

These methods are often sensitive to initial conditions and cannot guarantee global optimality, unlike the family of Branch and Bound (BnB) methods which operate in the rotation search space \cite{BazinBnB, HartleyBnB, ParraBnB}. These methods guarantee global optimality, but cannot be considered real-time algorithms.
Real-time rotation estimation is enabled in \cite{JooBnB} using the BnB method by maximising the consensus set of inliers over the search space of rotation. Surface normals are discretised on an equi-rectangular plane to generate the Extended Gaussian Image (EGI) \cite{HornEGI}, from which the BnB approach is used to estimate the MF.

Recently \cite{MNMA} proposed a novel and efficient cost function of Multiple Normal vectors and Multiple MF Axes (MNMA) for MF estimation. The cost function makes use of the vector dot and cross products between the scene surface normals and the axes of the MF leading to an efficient, accurate and real-time algorithm.

All of the methods considered here have used surface normals calculated from depth data from RGB-D images, rather than directly estimating them from an RGB input. Estimating surface normals has traditionally been computationally intensive, producing unreliable results. 
We make use of the recent advances in fast and accurate surface normal estimation \cite{bae2024dsine}, and build upon \cite{MNMA} to enable fast rotation estimation directly from RGB images. Unlike the previous work in this field, our method also captures per-frame estimates of rotational uncertainty.
This enables us to perform uncertainty-aware robust multi-frame optimisation leading to temporal consistency between frames.



\section{Method}


Given a monocular video, our goal is to estimate the per-frame camera rotation relative to the world coordinates. We begin by assuming that the scene satisfies the MW assumption~\cite{MWVP_2000_MW}, which is valid for a wide range of indoor/outdoor scenes. Note that it is straightforward to extend our approach to other world assumptions (e.g. Mixture of Manhattan Worlds~\cite{StraubMMF}, Atlanta World~\cite{schindler2004atlanta}, and Hong Kong World~\cite{li2023hong}), given that the principal directions in the world coordinates are known \textit{a priori}. 

Our method is named \textbf{U-ARE-ME} (Uncertainty-Aware Rotation Estimation in Manhattan Environments), as it can be used to complement or replace the I-M-U sensors. We leverage the recent advances in single-image surface normal estimation and propose to infer the camera rotation by aligning the predicted normals to the world assumption. In Sec.~\ref{sec:method1} we introduce a new uncertainty-weighted optimisation objective and show how the uncertainty can be learned from data. For real-time video applications, it is important to ensure temporal consistency in the predictions. We explain in Sec.~\ref{sec:method2} the factor graph formulation required to achieve this.

\subsection{Uncertainty-aware rotation estimation from a single image}
\label{sec:method1}

\begin{figure}[t]
\centering
\includegraphics[width=1.0\linewidth]{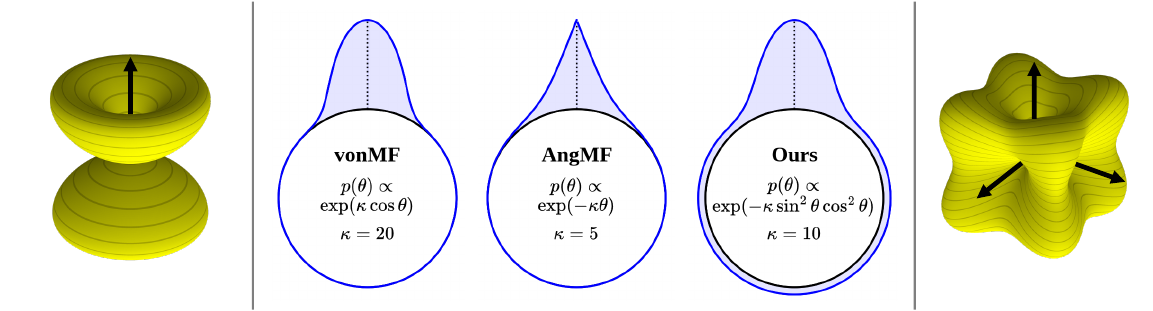}
\caption{\textbf{(left)} This figure visualises the cost function defined by a single normal vector. The cost is minimised when the (rotated) principal axes are parallel or vertical. \textbf{(middle)} Here we compare the shape of different probability distributions defined over a unit sphere. \textbf{(right)} This figure visualises the cost function defined by three mutually orthogonal Manhattan axes.}
\label{fig:method1}
\end{figure}

Suppose that a surface normal vector $\mathbf{n}_i \in \textit{S}^2$ corresponding to the $i$-th pixel is aligned with one of the principal directions. Then, for any Manhattan axis $\mathbf{r} \in \{ \pm X, \pm Y, \pm Z \}$, the angle $\theta = \cos^{-1} (\mathbf{n}_i \cdot \mathbf{r})$ should be $\{ 0^\circ, 90^\circ, 180^\circ \}$. To this end, Zhang et al.~\cite{MNMA} introduced a cost function $E(\mathbf{r}|\mathbf{n}_i)=\sin^2 \theta \cos^2 \theta$, which is visualised in Fig.~\ref{fig:method1}. We modify this cost by multiplying it by some confidence measure $\kappa$.

\begin{equation}
\label{eqn:method-cost}
E(\mathbf{r}|\mathbf{n}_i, \kappa_i) = \kappa_i \sin^2 \theta \cos^2 \theta
\end{equation}

To learn $\kappa$ in a data-driven manner, we define the following training loss:

\begin{equation}
\label{eqn:method-loss}
\mathcal{L}(\mathbf{n}^{gt}_i|\mathbf{n}_i,\kappa_i) 
= C(\kappa_i) + \kappa_i \sin^2 \theta \cos^2 \theta
\end{equation}

\noindent
where $\mathbf{n}^{gt}_i$ is the ground truth and $\theta$ is the angular error of the predicted normal $\mathbf{n}_i$. $C(\kappa)$ should be a monotonically decreasing function of $\kappa$ to prevent the model from estimating $\kappa_i = 0$ for every pixel. Another thing to note is that the second term should be defined only for $0^\circ \leq \theta < 45^\circ$. Otherwise, the loss can be minimised by \textit{increasing the error}. To satisfy such constraints, we assume that the surface normal probability distribution can be parameterised as follows:

\begin{equation}
\label{eqn:method-pdf}
\begin{aligned}
p(\mathbf{n}^{gt}_i|\mathbf{n}_i,\kappa_i) 
&= 
\begin{cases}
D(\kappa_i) \exp(- \kappa_i \sin^2 \theta \cos^2 \theta)
\;\;\; \text{when} \;\;\;
\theta < \frac{\pi}{4}\\
D(\kappa_i) \exp(- \frac{\kappa_i}{4})
\;\;\; \text{when} \;\;\;
\theta >= \frac{\pi}{4}
\end{cases} \\
\text{where}\;\;\; C(\kappa_i) &= - \log D(\kappa_i).
\end{aligned}
\end{equation}

Then, $\mathcal{L}(\mathbf{n}^{gt}_i|\mathbf{n}_i,\kappa_i)$ can be interpreted as the negative log-likelihood of the above distribution. $D(\kappa)$ is a monotonically increasing function of $\kappa$ as the distribution should be normalised. During training, the network is encouraged to increase the value of $\kappa$ for the pixels with lower error, thereby encoding confidence in the prediction. In Fig.~\ref{fig:method1}, we provide a visual comparison of the proposed distribution against the von Mises-Fisher distribution~\cite{OTHER_1993_vMF} and the Angular vonMF distribution~\cite{SNfromRGB_2021_EESNU}. As the analytic form for $D(\kappa)$ could not be found, we obtain the values for $\kappa \in [0, 10^5]$ and fit them using natural cubic splines. We use a lightweight convolutional encoder-decoder architecture~\cite{densedepth} and use the training data of \cite{bae2024dsine}. See Appendix \ref{sec:app_training} for additional detail regarding the network training.

Using the formulation of the problem in \cite{MNMA}, Eq. \eqref{eqn:method-cost} can be given in terms of a rotation matrix $\mathbf{R}\in SO(3)$ that gives the rotation from the camera to the world frame. We then use Levenberg-Marquardt (LM) optimisation to minimise the cost function, rewriting it in terms of the corresponding residual function $f(\mathbf{R})$ to obtain the optimal rotation $\mathbf{R^*}$ (see \cite{MNMA} for a full derivation).
\begin{align}
    \mathbf{R^*} &= \argmin_{\mathbf{R}} E(\mathbf{R}|\mathbf{n},\kappa) \\
                 &= \argmin_{\mathbf{R}} f(\mathbf{R})^\intercal f(\mathbf{R}).
\end{align}

From \cite{MNMA}, the analytical Jacobian of the residual function, $J_f(\mathbf{R})$, is given with respect to a change in $\mathbf{R}$ i.e. $\text{exp}(\Delta\phi)\mathbf{R}$, where $\Delta\phi$ refers to the Lie algebra of a change in rotation about $\mathbf{R}$ (and therefore about $\mathbf{r}$). This allows us to acquire an approximation for the covariance $\Sigma^{R}$ as:
\begin{align}
    \label{eq:jac}
    J_f(\mathbf{R}) &= \frac{\partial f(R)}{\partial\Delta\phi}\\
    \label{eq:sigma}
    \Sigma^{R} = H^{-1} &\approx (J_f(\mathbf{R})^\intercal J_f(\mathbf{R}))^{-1} 
\end{align}
where $H$ is the hessian matrix approximation from the second order optimisation.

Having obtained the optimal MW rotation $\mathbf{R}_{\text{mw}}=\mathbf{R^*}^\intercal$ and its uncertainty, this estimate can then be used for further downstream tasks. This could be for bootstrapping visual odometry systems, correcting drift in inertial pipelines, rectifying images for CNNs, which are sensitive to image rotation, as well as camera calibration to name a few. In the following section we address one such extension, that being to estimate camera rotation across a sequence of images to achieve a coherent trajectory.

\subsection{Multi-frame rotation estimation for temporal consistency}
\label{sec:method2}

\begin{figure}[t]
    \centering
    \includegraphics[width=1.0\linewidth]{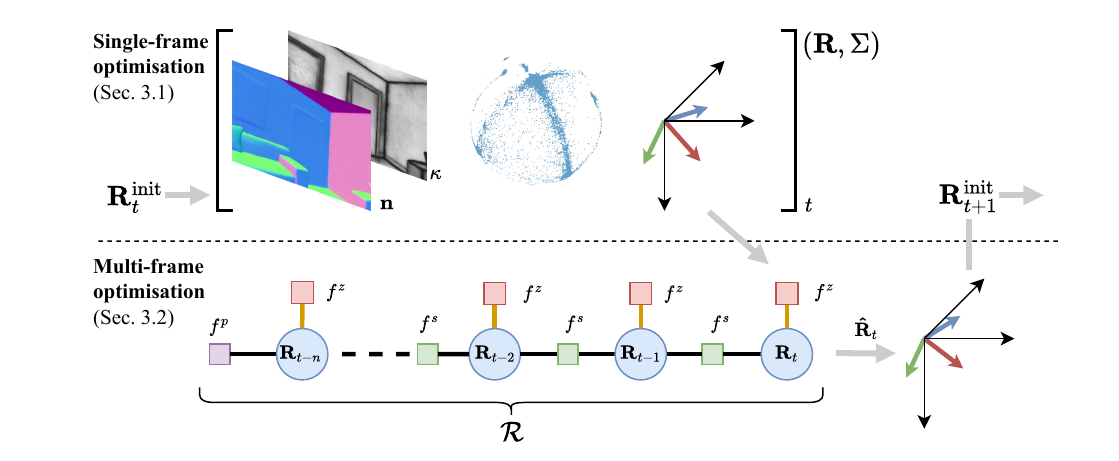}
    \caption{The multi-frame optimisation process. Single-frame rotation and covariance estimates are used to initialise a sliding window factor graph in order to provide temporal consistency between frames and reject outlier measurements. Robust factors are shown along orange edges on measurements. The latest frame is then used to initialise the rotation estimate for the next frame.}
    \label{fig:multiframe}
\end{figure}

When estimating single-frame rotation using the above method, the rotation relative to the closest minimum of the cost function is found. The problem with applying this method consecutively to a sequence of images is that there will be no temporal consistency between rotations, and therefore when rotating around any single axis, there is an inherent $90^{o}$ ambiguity which will always default to the closest rotation to identity. It is simple to naively initialise the current rotation $\mathbf{R}_t$, with the optimised rotation from the previous frame $\mathbf{R}_{t-1}$ which removes this ambiguity and also increases convergence of optimisation. However, two main problems arise from this initialisation. Firstly, for frames that are less Manhattan, we have no way of loosening the MW assumption and therefore a single frame's rotation is only dependent on its own (potentially poorly defined) cost function . Secondly, not all frames in a sequence may optimise to a consistent minimum and may give erroneous rotations that initialise subsequent frames with a bad orientation, potentially poisoning the rest of the sequence. In the following section, we address these issues in order to extend rotation estimation to sequences of images.

\subsubsection{Sliding window optimisation}
To tackle both of these issues, we implement a sliding window optimisation that jointly considers previously estimated frame rotations. This sliding window acts as a support for the current rotation estimate in order to provide consistency along the sequence. To model this non-linear joint optimisation, we employ a simple factor graph that consists of factor nodes $f \in \mathcal{F}$ and variables $\mathcal{X} = \{x_t, \ldots, x_{t-n}\}  $, where $t$ refers to the latest variable and $n$ defines the length of the sliding window. The joint distribution $p(\mathcal{X})$ is represented as a product of all factors in the graph which, when considering a Gaussian factor graph, can be solved by the following minimisation:

\begin{align}
    \hat{\mathcal{X}} = \argmin_{\mathcal{X}} \sum_i{||h_i(\mathcal{X}_i) - z_i||^2_{\Sigma_i}}
\end{align}
where $\mathcal{X}_i$ represents the clique of variables corresponding to the factor $f_i$, $h_i(\cdot)$ represents a function that predicts a measurement based on the state of input variables, and $z_i$ represents some observed measurement. For further details on state estimation with factor graphs, see Dellaert et. al's book on the subject \cite{Dellaert2017}.

For our problem, each variable $x_i \in \mathcal{X}$ represents an $SO(3)$ rotation $\mathbf{R}_i \in \mathcal{R}$, and 3 types of factors exist to constrain the problem. The first factor, $f^{z}(\mathbf{R}_i)$, is a prior factor on each variable that is set to the rotation estimated from the single-frame estimation problem, $\mathbf{Z}_i \in SO(3)$. The second factor, $f^{s}(\mathbf{R}_i,\mathbf{R}_j)$, is a smoothness factor which enforces that adjacent frames should have a similar rotation i.e. their difference should be the identity rotation, $\mathbf{I}_3$. Finally, another prior factor $f^{p}(\mathbf{R}_i)$ is added to the oldest frame in the sliding window which represents the marginalised state of the oldest variable in the previous sliding window optimisation, $\mathbf{R}_p$. This multi-frame factor graph is shown pictorially in Figure \ref{fig:multiframe}. The minimisation for our problem is therefore given by:

\begin{align} \label{eq:mf_minso3}
    \hat{\mathcal{R}} = \argmin_{\mathcal{R}} \sum_{(i,j)\in\mathcal{F}^s}{||\mathbf{R}_i^{-1}\mathbf{R}_j \ominus \mathbf{I}_3||^2_{\Sigma^s_i}} + \sum_{i\in\mathcal{F}^z}{||\mathbf{R}_i \ominus \mathbf{Z}_i||^2_{\Sigma^z_i}} + \sum_{i\in\mathcal{F}^p}{||\mathbf{R}_i \ominus \mathbf{R}_p||^2_{\Sigma^p_i}}
\end{align}
where $\ominus$ denotes the vector increment (defined on the tangent space of the right hand variable) between two rotations via the logarithmic map. See \cite{sola2018} for more details on the subject of lie theory when applied to 3D rotations. For ease of implementation, we use the popular sensor fusion library GTSAM \cite{gtsam} to optimise the multi-frame factor graph based on the definitions above.

Once $\hat{\mathbf{\mathcal{R}}}$ has been computed, the latest frame's rotation can then be fed back into the single frame optimisation as initialisation for the subsequent frame i.e. $\mathbf{R}^{\text{init}}_{t+1} = \mathbf{\hat{R}}_t \in \hat{\mathcal{R}}$

\subsubsection{Robust estimation}
In order to reduce the influence of outlier measurements due to dropped frames, poor normal predictions, and incorrect local minima from the single frame optimisation process, we also apply robust factors to all prior measurement factors $f^{z}$. We leverage the Huber cost function which represents a Gaussian energy for small residuals, but transitions to a linear function for large residuals. This effectively dampens the influence of measurements that grossly disagree with the smoothness model between variables. However, by maintaining these measurements in the factor graph, a genuine large change in rotation will be properly estimated after a few frames of consistent measurements.    

The result of this robustified multi-frame optimisation is that high frequency noise in the single frame rotation estimates is smoothed out, whilst outlier measurements are repressed from poisoning the rotation estimates of subsequent frames.

\subsubsection{Incorporating uncertainty}
To address the issue of non-Manhattan frames, the covariance estimate from Eq. \eqref{eq:sigma} can be directly applied to each measurement factor ($\Sigma^z$ in Eq. \eqref{eq:mf_minso3}), since Eq. \eqref{eq:jac} is defined around the global MF upon which we are optimising. This means that frames which exhibit strong Manhattan normals will have a greater influence on the result, stabilising rotation estimates for non-Manhattan frames. For example, if a frame only sees normals $\mathbf{n}$ in one dominant direction, $\mathbf{r}_i$, then the single-frame optimisation is free to rotate around this vector. Since $\lim_{\mathbf{n} \rightarrow \mathbf{r}_i} \sigma^2_{\mathbf{r}_i} = \infty$, whilst $\lim_{\mathbf{n} \rightarrow \mathbf{r}_i} \sigma^2_{\mathbf{r}_i^{\perp}} = 0.5$, any rotation around $\mathbf{r}_i$ will be effectively ignored in the multi-frame optimisation, whilst keeping the information for the perpendicular axes of rotation. Therefore even though we are using the MW assumption, we are not strictly bound to it.

For the smoothness factors, the covariance $\Sigma_s$ is set to an isotropic covariance $\lambda\mathbf{I}_3$, where $\lambda$ is a tuning parameter that defines how strongly the smoothness constraint is enforced. A high value will ensure rotations are close to each other, whilst a low value will trust the measurement rotations more. The covariance for the prior measurement $\Sigma^p$, is automatically defined by the marginalisation of the last variable $\mathbf{R}_{t-n}$ in the previous iterations window.


\section{Experiments}
\label{sec:exp}
In Sec.~\ref{sec:exp1}, we evaluate the accuracy and robustness of U--ARE--ME and make a comparison to the existing approaches. In Sec.~\ref{sec:exp2}, we demonstrate the versatility of our approach by introducing potential applications: up-vector estimation and ground segmentation.

\subsection{Rotation estimation over sequences}
\label{sec:exp1}



\subsubsection{Dataset and evaluation protocol}

Following previous methods~\cite{e-graph, compass, guo2019robust}, we evaluate our method on ICL-NUIM~\cite{ICL-NUIM} and TUM RGB-D~\cite{TUM-RGBD} which cover synthetic and real indoor scenes, respectively. To assess the performance in challenging real-world scenarios, we also evaluate the performance on ScanNet~\cite{scannet}. ScanNet images have a significant amount of noise and blur as they were captured using hand-held cameras in poorly lit environments. The scenes are also cluttered with many objects that violate the Manhattan World assumption. We do not discard such scenes and evaluate on all test sequences to highlight the robustness of our method.

To measure accuracy, each frame is fed into the algorithm sequentially and the estimated rotation is recorded after each frame. The rotations are then aligned with the relevant ground truth so that methods that do not estimate a specific world alignment can also be compared with the MW-based methods. The metric used for accuracy is the average rotation error (ARE) and is given by
\begin{equation}
    \text{ARE} = \text{cos}^{-1}\left(\frac{\text{tr}({\mathbf{R}}_{\text{gt}}^{-1} \hat{\mathbf{R}})-1}{2}\right)
\end{equation}
where $\hat{\mathbf{R}}$ is the rotation estimate and ${\mathbf{R}_\text{gt}}$ is the ground truth.


\subsubsection{Baseline methods} We compare our approach to various monocular rotation estimation methods. We first choose OLRE~\cite{olre} and ROVE~\cite{Lee-ROVE} which extract and align vanishing points. We also compare against RGB-D approaches that align the surface normal vectors extracted from the depth measurements and replace with our predicted normals to assess their performance in a monocular setup. Furthermore, we also draw direct comparisons with the popular monocular SLAM system ORB-SLAM \cite{mur2015orb}. Whilst this is not a single image rotation estimation method, it is one of the state-of-the-art methods for obtaining accurate real-time odometry and will provide a challenging benchmark from which to draw conclusions of our work. Lastly, several RGB-D methods are also used in our comparisons so that we can give context on how accurate our system is compared to methods that require a depth map. These are: GOME \cite{JooBnB}, Compass \cite{compass} and  E-Graph \cite{e-graph}.




\subsubsection{Results from ICL-NUIM and TUM RGB-D}

The overall results for both the ICL-NUIM and TUM RGB-D datasets are shown in Tab. \ref{tab:results_icltum}. U--ARE--ME can accurately estimate a valid trajectory of rotations in 14/15 sequences and has on average the best accuracy of the non-SLAM based monocular methods. Whilst RTMF, RMFE and ES sometimes show the same accuracy as the proposed method, the lack of temporal consistency means that they often shift into a globally inconsistent MF and in many of the sequences show large errors. Comparing to the RGB-D methods, the proposed method is often better than GOME despite only using RGB images, and whilst we are worse than Compass and E-Graph for the synthetic ICL-NUIM sequences (which have perfect depth), we achieve comparable accuracy in some real-world TUM sequences.

\begin{table}[t]\centering
\caption{\textbf{(top)} Quantitative evaluation on ICL-NUIM and TUM RGB-D [deg]. The best RGB method is \textbf{bold} and the second-best is \underline{underlined}. \textbf{(bottom)} Our approach, compared to ORB-SLAM, is more robust to inaccuracy in camera intrinsics. We simulate changes in focal length and principal point by cropping/shifting the input image.}
\label{tab:results_icltum}
\scriptsize
\begin{tabular}{lcccccccccccc}
\toprule
\multirow{2}{*}{\textbf{Sequences}}& &\multicolumn{7}{c}{\textbf{RGB Methods}} & &\multicolumn{3}{c}{\textbf{RGB-D methods}} \\
& &\textbf{Ours} &RMFE* &RTMF* &ES* &OVPD &REOV &ORB & &GOME &Compass &E-Graph \\
\cmidrule{1-1}
\cmidrule{3-9}
\cmidrule{11-13}
Office 0 & &4.99 &4.99 &\underline{4.97} &5.21 &6.71 &29.11 &\textbf{0.60} & &5.12 &0.37 &0.11 \\
Office 1 & &\textbf{3.87} &89.18 &44.59 &\underline{3.90} & $\times$ &34.98 &$\times$ & &$\times$ &0.37 &0.22 \\
Office 2 & &2.38 &3.35 &\underline{2.36} &41.99 &10.91 &60.54 &\textbf{0.69} & &6.67 &0.38 &0.39 \\
Office 3 & &\underline{2.72} &2.84 &41.98 &2.87 &3.41 &10.67 &\textbf{2.53} & &5.57 &0.38 &0.24 \\
Living 0 & &8.43 &8.36 &\underline{8.25} &11.53 &$\times$ &$\times$ &\textbf{0.35} & &$\times$ &0.31 &0.44 \\
Living 1 & &\textbf{3.58} &91.95 &3.81 &14.04 &\underline{3.72} &26.74 &$\times$ & &8.56 &0.38 &0.24 \\
Living 2 & &\underline{2.39} &2.45 &2.50 &2.44 &4.21 &39.71 &\textbf{0.57} & &8.15 &0.34 &0.36 \\
Living 3 & &\underline{5.38} &5.64 &5.58 &57.62 &$\times$ &$\times$ &\textbf{0.84} & &$\times$ &0.35 &0.36 \\
\cmidrule{1-1}
Struc notex & &\underline{4.61} &\textbf{4.55} &4.94 &7.96 &11.22 &$\times$ &$\times$ & &4.07 &1.96 &4.46 \\
Struc tex & &\underline{3.03} &3.10 &3.18 &3.14 &8.21 &13.73 &\textbf{0.37} & &4.71 &2.92 &0.60 \\
Large cabinet & &4.54 &\underline{4.30} &4.60 &5.20 &38.12 &28.41 &\textbf{1.13} & &3.74 &2.04 &1.45 \\
Cabinet & &\textbf{5.41} &40.15 &\underline{6.27} &70.39 &$\times$ &$\times$ &$\times$ & &2.59 &2.48 &2.47 \\
Long office & &\textbf{5.62} &\underline{5.78} &5.98 &46.74 &$\times$ &$\times$ &7.86 & &$\times$ &1.75 &- \\
Nostruc notex & &\textbf{6.93} &54.46 &\underline{27.52} &30.77 &$\times$ &$\times$ &$\times$ & &$\times$ &$\times$ &- \\
Nostruc tex & &28.94 &24.16 &63.62 &\textbf{11.58} &46.18 &\underline{16.45} &17.42 & &$\times$ &$\times$ &- \\
\bottomrule
\multicolumn{13}{l}{* Reimplemented using our predicted normals} \\
\\
\\
\end{tabular}

\begin{tabular}{lcccclccc}
\toprule
\textbf{Crop} & &Ours & ORB-SLAM & \quad\quad & \textbf{Shift} & &Ours & ORB-SLAM \\
\cmidrule{1-4}
\cmidrule{6-9}
Original & &2.39 & \textbf{0.57} & \quad  & original & & 2.39 & \textbf{0.57}\\
5\% crop & &\textbf{2.30} & 2.40 & \quad & 5\% shift & & \textbf{2.31} & 6.36\\
10\% crop& &\textbf{2.28} & 5.13 & \quad & 10\% shift & & \textbf{3.97} & 14.63\\
15\% crop & &\textbf{2.31} & 7.45 & \quad & 15\% shift & & \textbf{6.03} & 22.32\\
20\% crop & &\textbf{2.39} & X & \quad & 20\% shift & & \textbf{8.26}  & X\\
\bottomrule
\end{tabular}

\end{table}


Comparing to ORB-SLAM, for the ICL-NUIM sequences in which ORB-SLAM does not fail, we find that ORB-SLAM is significantly better than all the RGB methods and is even on par with the RGB-D methods. However, ORB-SLAM may lose track or be unable to initialise in texture-less scenes, meaning that it sometimes fails to register many valid rotations. ORB-SLAM (and indeed most other SLAM methods) also need finely calibrated camera intrinsics, and are therefore often not suitable for running on \textit{in-the-wild} sequences which may be subject to distortion and cropping (something we are capable of handling with U--ARE--ME). To demonstrate this point, Tab. \ref{tab:results_icltum} (bottom) shows the accuracy comparison on the ICL-NUIM living room 2 sequence, but using increasingly deformed images with the same camera intrinsics on both U--ARE--ME and ORB-SLAM. Since we do not need specific camera intrinsics, we are extremely robust to both image cropping and image shifting. Since the normal network has been trained on \tildetext$60^o$ fov images, we actually see a minor improvement at 90\% cropping as the ICL-NUIM images have a slightly wider than ideal $67^o$ fov. 



\subsubsection{Results from ScanNet}

\begin{figure}[t]
\centering
\includegraphics[width=1.0\linewidth]{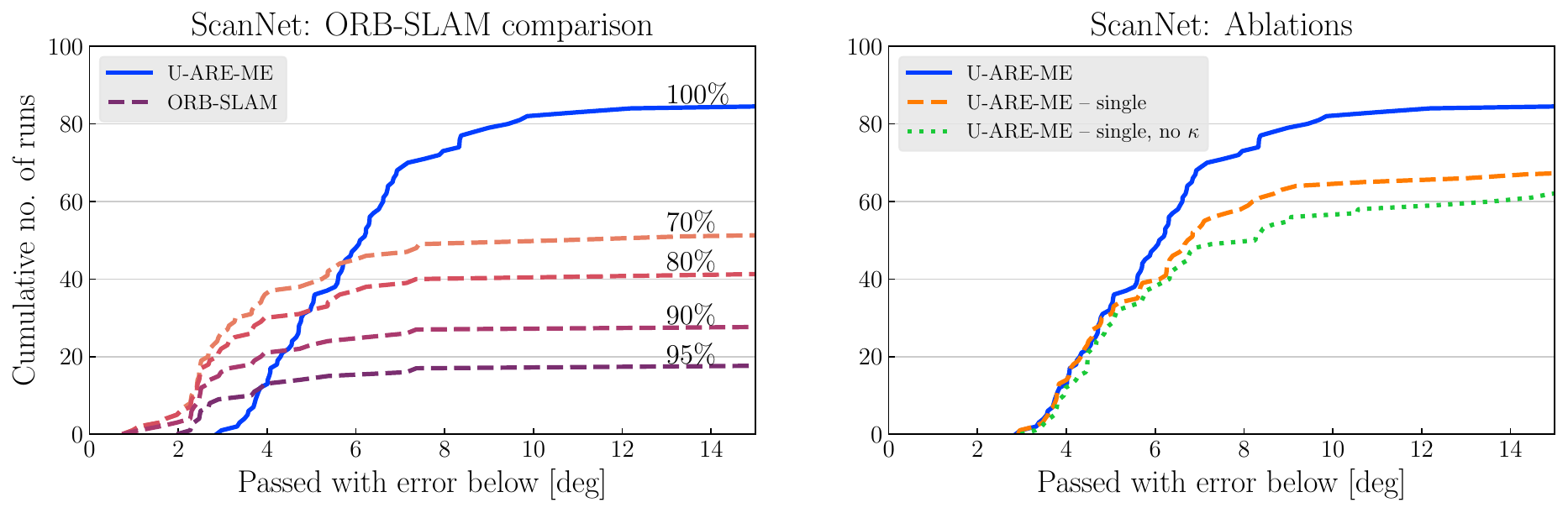}
\caption{\textbf{(left)} U--ARE--ME and ORB-SLAM accuracy comparison on 100 sequences from the ScanNet dataset. Solid line shows cumulative number of runs below a certain accuracy threshold for U--ARE--ME and dashed lines are for ORB-SLAM. Percentage pass rate is shown, whereby at least X\% of frames per sequence must contain a valid rotation estimate (this includes any initialisation and loss of tracking). \textbf{(right)} Ablation study experiments. The blue line shows the results of the full pipeline. 'single' means that multi-frame optimisation is disabled and 'no $\kappa$' means that the uncertainty weighting in the cost function is removed.}
\label{fig:results_scannet}
\end{figure}

The ScanNet suite provides a large set of real-world sequences from which we can draw better conclusions about the generalisability of our system. We therefore further test U--ARE--ME on all 100 test sequences and compare to the most accurate RGB method from the previous section, ORB-SLAM. We then perform an ablation study on each of the proposed features of U--ARE--ME to demonstrate the benefit of each one.

Fig. \ref{fig:results_scannet} (left) shows the results from the comparison with ORB-SLAM. Since ORB-SLAM is a multi-view system and therefore will never produce rotation estimates for every single frame, we show the results of ORB-SLAM at different success rates e.g. 70\% defines that at least 70\% of frames per sequence need valid rotation estimates (the missing frames being from either initialisation or loss of tracking). In this regard, we allow ORB-SLAM to lose tracking and do not consider this an outright failure. The results show that U--ARE--ME has a much higher robustness to the ScanNet sequences and will estimate rotations with an accuracy $<10^o$ in 80\% of the sequences. Comparing this to ORB-SLAM with a 95\% frame threshold which only successfully achieves $<10^o$ in 17\% of the sequences. For higher accuracy $<3^o$, we find that ORB-SLAM is more capable in some sequences (10\%) whereas the proposed method only achieves at best $3^o$. We argue that this is primarily caused by the accuracy of the surface normal predictions, e.g. the state of the art \cite{bae2024dsine} reports a mean normal error of 16.2$^o$ on ScanNet, and therefore as normal predictions improve so should our results.

The ablation study Fig. \ref{fig:results_scannet} (right), shows that the overall reliability of the system improves as firstly, the uncertainty weighted normals reject non-Manhattan pixels within the image, and then more so second, the multi-frame optimisation provides a consistent global MF which anchors the solution across the sequence. We also see a slight improvement in the accuracy of the rotations when discounting uncertain pixels.

\subsection{Applications}
\label{sec:exp2}

\subsubsection{Up-vector Estimation}
Estimating the `upward' direction in a scene is important in applications that require knowledge of the scene orientation. Recent neural network approaches have shown success in estimating camera parameters such as the up-vector. CTRL-C \cite{CTRL-C}  proposes an end-to-end transformer approach to combine detected vanishing points with learned features. Perspective Fields (PF) \cite{PerspectiveFields} predicts the per-pixel information about the camera parameters, and demonstrated the use of the up-vector for AR effects such as compositing rainfall and 3D objects into the scene.

We compare our method against these baselines on the ICL-NUIM dataset and report the angular difference of the up-vector in Table \ref{tab:upvector}. As the baseline methods operate on single RGB images, we perform single frame rotation estimation in our method for a fair comparison. Our method outperforms the baselines which were trained on feature-rich image datasets, and is able to more accurately estimate camera rotation in the relatively texture-less scenes from the ICL-NUIM dataset.


\begin{table}[t]
\caption{Accuracy of the estimated up-vector [deg]. Note that PF~\cite{PerspectiveFields} and CTRL-C~\cite{CTRL-C} were trained specifically to estimate the up-vector.}
\label{tab:upvector}
\centering
\scriptsize
\begin{tabular}{@{}lcccclccc@{}}
\toprule
\textbf{Sequence} & U--ARE--ME & PF & CTRL-C &\quad&
\textbf{Sequence} & U--ARE--ME & PF & CTRL-C \\ 
\cmidrule{1-4}
\cmidrule{6-9}
Living 0 &\textbf{7.53} & 9.90  & 12.14 &\quad& Office 0 &\textbf{3.95} & 5.52  & 18.87 \\
Living 1 &\textbf{2.67} & 3.17  & 13.74 &\quad& Office 1 &\textbf{3.50} & 4.65  & 22.32 \\
Living 2 &\textbf{1.77} & 4.67  & 9.71  &\quad& Office 2 &\textbf{1.61} & 3.80  & 15.95 \\
Living 3 &\textbf{4.01} & 10.84 & 7.67  &\quad& Office 3 &\textbf{2.13} & 3.95  & 18.47 \\
\bottomrule
\end{tabular}%
\end{table}

\begin{figure}[t]
  \centering
  \begin{subfigure}{0.2\linewidth}
    \includegraphics[width=\linewidth]{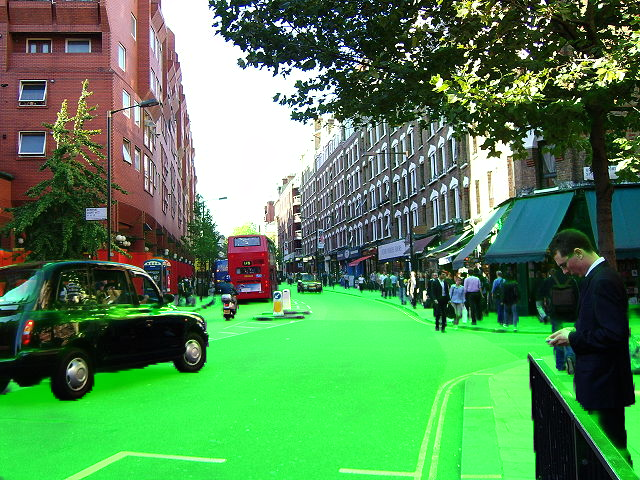}
    \caption{London}
    \label{fig:short-a}
  \end{subfigure}
  \begin{subfigure}{0.2\linewidth}
    \includegraphics[width=\linewidth]{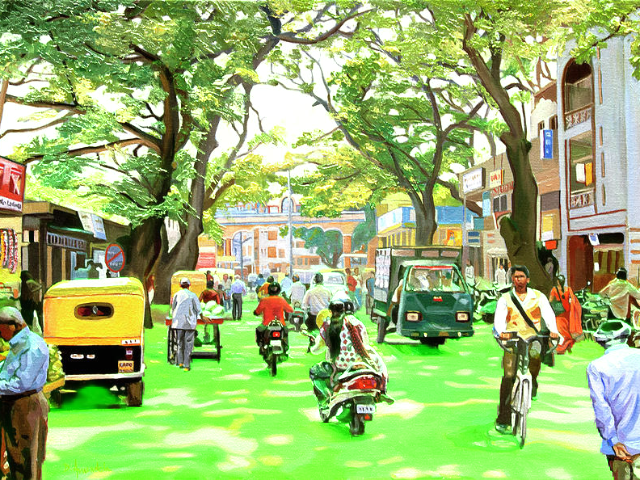}
    \caption{Mumbai}
    \label{fig:short-b}
  \end{subfigure}
  \begin{subfigure}{0.2\linewidth}
    \includegraphics[width=\linewidth]{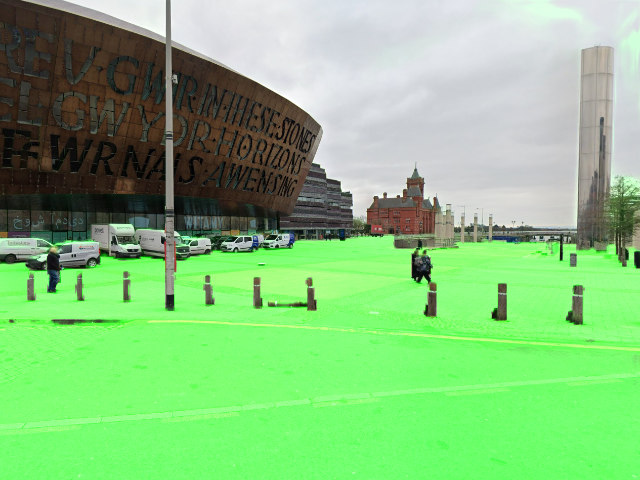}
    \caption{Cardiff}
    \label{fig:short-c}
  \end{subfigure}
  \begin{subfigure}{0.2\linewidth}
    \includegraphics[width=\linewidth]{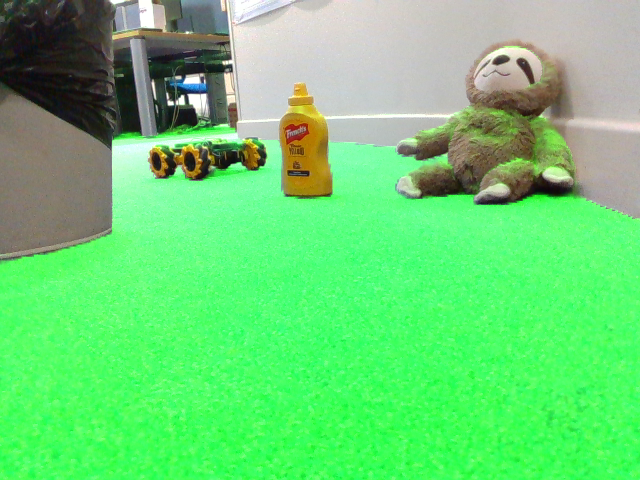}
    \caption{Office}
    \label{fig:short-d}
  \end{subfigure}
  \caption{Even in non-Manhattan scenes, our framework can optimise the rotation such that the global up direction is aligned with the surface normal of the ground plane. It can thus be used to accurately segment the ground plane, which can be useful for robotics applications.}
  \label{fig:ground_seg}
\end{figure}

\subsubsection{Horizon estimation in a Non-inertial Reference Frame}
We are also motivated by the limitations of using Inertial Measurement Units (IMUs) when operating within a non-inertial reference frame. We perform multi-frame rotation estimation for a sequence of RGB images taken inside an accelerating London bus, and perform a comparison against IMU data.
For each frame we compute the ARE between the estimated rotation and the initial rotation, and plot the evolution over time in Fig. \ref{fig:IMU_horizon}.

The camera is stationary with respect to the bus throughout the sequence and so there should be no relative rotation.
The Intel Realsense D435i camera was used to take synchronised RGB and IMU measurements which were integrated using the complementary filter provided by the Intel Realsense SDK \cite{realsense}.

As the bus harshly turns our visual-only approach maintains a steady rotation estimate, while the IMU suffers from strong accelerations causing a large error in its measured rotation. This is seen in Fig. \ref{fig:IMU_horizon} where the horizon lines for each method have been calculated using the up-vector derived from the estimated rotation. We envisage that U--ARE--ME can be used to complement traditional IMU-based applications in non-inertial reference frames.

\subsubsection{Ground Segmentation}
The up-vector can be used to perform real-time ground segmentation from RGB images, which is an important pre-processing step in many applications including robotics \cite{GroundNet}, autonomous driving \cite{groundsegdriving}, and 3D object tracking for augmented reality \cite{augreality}, where it can be used to seamlessly integrate virtual objects with the real-world environment.

As U--ARE--ME produces per-pixel surface normal estimates, we can directly use the result of the rotation estimation to segment areas of an input image corresponding to the ground, assuming that this is aligned with the world up-vector. Our method can be applied to real-world indoor and outdoor images to segment the ground even when the scene is non-Manhattan, as seen in Fig. \ref{fig:ground_seg}.

\section{Conclusion}

Motivated by the need to provide extrinsic rotation estimates from \textit{in-the-wild} images and sequences, we have presented U-ARE-ME, an accurate and robust camera rotation estimator that operates on uncalibrated RGB images. The system is capable of outputting estimates for single images and has also been extended to reliably handle multi-frame scenarios. An extensive evaluation has been performed and it has been shown that our method is capable of providing globally consistent multi-frame rotation estimates which rivals the performance of similar methods that leverage accurate depth maps -- all whilst remaining real-time. By accounting for the uncertainty and inherent ambiguity of the common MW assumption, we are also capable of providing accurate results on scenes that superficially appear to not contain structural regularities, and where other 2D feature-based methods often fail.

\section*{Acknowledgements}
This research has been supported by the EPSRC Prosperity Partnership Award with Dyson Technology Ltd.

\bibliographystyle{splncs04}
\bibliography{egbib}

\appendix
\chapter*{Appendix}
\section{Training details for surface normal estimator}
\label{sec:app_training}

Our surface normal estimator is trained on the meta-dataset introduced by \cite{bae2024dsine}. DSINE~\cite{bae2024dsine} uses per-pixel ray direction as input and thus requires camera intrinsics. We removed this dependency and warped the training images such that the principal point is at the center and the field-of-view is $60^\circ$. Nonetheless, U--ARE--ME generalises well to images taken with different intrinsics (e.g. the video game sequence in the attached video has field-of-view of $86^\circ$). DSINE~\cite{bae2024dsine} also proposed to improve the piece-wise smoothness and crispness of the prediction by recasting surface normal estimation as iterative rotation estimation. We also remove this iterative process and hence improve the efficiency to give real-time estimates. While this degrades the quality of the surface normal prediction, the rotation estimates from our framework stay robust. U--ARE--ME robustly fuses the per-pixel predictions by weighting them with a confidence $\kappa$ and is thus robust to mild inaccuracies in the surface normal prediction. 

Our network estimates two quantities: the per-pixel surface normal vector $\mathbf{n}$ and the corresponding confidence $\kappa$. $\mathbf{n}$ is supervised with the angular loss, and $\kappa$ with the negative log-likelihood defined in Eq. 2 (in the main manuscript). All other training protocols (e.g. batch size, number of epochs, data augmentation, optimiser, and learning-rate schedulers) are the same as \cite{bae2024dsine}. The training only takes 9 hours on a single NVIDIA 4090 GPU. After training, $\kappa$ is capped at $100$ to prevent the over-confident predictions from dominating the optimisation.

\section{Demo video}
\label{sec:app_video}

We encourage the reader to view the accompanying video at https://callum-rhodes.github.io/U-ARE-ME for a visual overview of our method.

For the reader's convenience, the timestamps of each section in the video are summarised in Table \ref{tab:timestamps}. We provide additional detailed explanations for each section below.
\begin{table}[h]
\caption{Timestamps for the contents of the demo video}
\label{tab:timestamps}
\centering
\begin{tabular}{llc}
\toprule
\textbf{Section}                   &                 & \textbf{Timestamp} \\ \midrule
U--ARE--ME Demo         & ICL-NUIM                   & 0:13      \\
         & TUM-RGBD                   & 0:30      \\
         & Tokyo walking sequence                   & 0:47      \\
         & Video game sequence                   & 1:05      \\
Robustness to Dropped Frames   &                & 1:29      \\
Applications & Non-inertial Reference Frame & 1:58      \\ 
 & Ground Segmentation          & 2:27  
\\ \midrule
\end{tabular}
\end{table}

\subsection{U--ARE--ME demo}
We first demonstrate the operation of U--ARE--ME on the ICL-NUIM and TUM-RGBD datasets discussed in our paper. The coordinate frame in the center of the video depicts the orientation of the global Manhattan frame.
Throughout the demonstration, the video cycles through various visualisations of the scene:
\begin{itemize}
    \item RGB image input to U--ARE--ME
    \item Predicted surface normals (using X--Y--Z to R--G--B colour mapping)
    \item Confidence of predicted surface normals (greyscale -- white represents high confidence)
\end{itemize}

\subsubsection{ICL-NUIM: living-room-2}
This synthetic scene shows a camera moving through a living room which is generally Manhattan in structure, but does contain some features which do not agree with a Manhattan assumption (e.g. curtains, lamps, and sofa cushions).
Of note is the textured wall painting at 0:13 which is predicted as having the same surface normal as the wall it is on, while the painting's frame is predicted to have high uncertainty normals. 
The curtains seen at 0:21 (a large non-Manhattan area) are also shown with high uncertainty normals as they have a irregular geometry. As U--ARE--ME is uncertainty-aware, it estimates rotation with a greater weighting on those surfaces that agree with the Manhattan assumption, e.g., the walls and floor (shown in white on the confidence images), whilst down-weighting the non-Manhattan features. 

\subsubsection{TUM-RGBD: fr-3 large cabinet}
This dataset contains videos captured with a real-world hand-held camera, exhibiting some pitch and roll with unsteady camera motion.
The normals of objects in the background with fine structures and lots of occlusion are harder to accurately predict, thus are estimated as having surface normals with higher uncertainty. As a result, our method can produce accurate estimates of rotation throughout the sequence by down-weighting such uncertain normals.

\subsubsection{Tokyo Walking Sequence}
We apply our method to an \textit{in-the-wild} video taken directly from YouTube \cite{YT_Tokyo_arxiv} showing a hand-held camera viewpoint walking through the streets of Tokyo. This is a challenging situation for rotation estimation as the camera intrinsics are not known.
The environment is dynamic, with many pedestrians walking in the scene, and our method produces reliable rotation estimation despite the small quantity of static Manhattan aligned objects and buildings.

\subsubsection{Video Game sequence}
In this example our method is shown to estimate rotation on a synthetic sequence from the video game Star Citizen \cite{YT_StarCitizen_arxiv}. Once again the camera intrinsics are not known, and this is a relatively non-Manhattan environment. During the sequence, the camera switches between 1st person and 3rd person (during which the player character takes up a significant portion of the screen) yet rotations remain consistent with the game world. U--ARE--ME takes this into consideration by estimating a high uncertainty on the player model.

\subsection{Robustness to dropped frames}
We compare U--ARE--ME operating \textbf{with} (bottom videos) and \textbf{without} (top videos) multi-frame optimisation, when black frames are randomly injected into the sequence, simulating dropped frames.

For example, at 01:50 three consecutive dropped frames are input to the non-multi-frame estimator (top) which causes the estimate to differ significantly from its previous estimate into a locally correct but globally inconsistent MF.
Using the proposed uncertainty-aware multi-frame optimisation however, U--ARE--ME is seen to produce temporally consistent rotation estimates, in the presence of dropped frames.

\subsection{Applications}
Finally we demonstrate some applications of U--ARE--ME as discussed in the paper.

\subsubsection{Operation in a Non-inertial Reference Frame}
We apply U--ARE--ME to a real-world video sequence captured from a bus with longitudinally and laterally accelerating motions. We estimate the global up-vector visualised as a horizon line (green). This is compared against the result obtained from an inertial measurement unit (IMU) only, using the Realsense camera's onboard state estimation.

The video also shows an outward view demonstrating the motion of the bus (left), the horizon line estimated by U--ARE--ME (middle), and the estimate based on IMU measurements (right).

\subsubsection{Ground Segmentation}
U--ARE--ME can keep track of the global up-vector, even in complex non-Manhattan sequences with high degrees of camera rotation without knowledge of the camera intrinsics. This can be useful in downstream tasks such as ground segmentation, by considering only the pixels with surface normals aligning with the global up-vector. 
For a video taken directly from YouTube \cite{YT_Parkour_arxiv} we use U--ARE--ME to estimate the up-vector and highlight the ground-aligned pixels in green.

\end{document}